\documentclass{article}

\usepackage{arxiv}

\usepackage[utf8]{inputenc} 
\usepackage[T1]{fontenc}    
\usepackage{url}            
\usepackage{booktabs}       
\usepackage{amsfonts}       
\usepackage{nicefrac}       
\usepackage{microtype}      
\usepackage{lipsum}		
\usepackage{graphicx}
\usepackage[square,numbers]{natbib}
\usepackage{doi}
\usepackage{amsmath}

\usepackage{algorithm, algpseudocode}
\usepackage{svg}

\title{SYNCHRONIZATION-BASED CLUSTERING ON THE UNIT 
HYPERSPHERE}

\author{Zinaid Kapić \\
	Faculty of Engineering \\
	University of Rijeka \\
    Rijeka \\
	\texttt{zkapic@uniri.hr} \\
	\And
	Aladin Crnkić \\
    Faculty of Technical Engineering \\
    University of Bihać \\
	Bihać \\
	\texttt{aladin.crnkic@unbi.ba} \\
    \And
	Goran Mauša \\
    Faculty of Engineering \\
	University of Rijeka \\
    Rijeka \\
	\texttt{goran.mausa@riteh.uniri.hr} \\
}

\renewcommand{\shorttitle}{SYNCHRONIZATION-BASED CLUSTERING ON THE UNIT 
HYPERSPHERE}

\hypersetup{
pdftitle={SYNCHRONIZATION-BASED CLUSTERING ON THE UNIT 
HYPERSPHERE},
pdfsubject={Computer Science},
pdfauthor={Zinaid Kapić, Aladin Crnkić, Goran Mauša},
pdfkeywords={ clustering, synchronization, unit-hypersphere},
}

\begin{document}
\maketitle

\begin{abstract}
	Clustering on the unit hypersphere is a fundamental problem in various fields, with applications ranging from gene expression analysis to text and image classification. Traditional clustering methods are not always suitable for unit sphere data, as they do not account for the geometric structure of the sphere. We introduce a novel algorithm for clustering data represented as points on the unit sphere $\mathbf{S}^{d-1}$. Our method is based on the $d$-dimensional generalized Kuramoto model. The effectiveness of the introduced method is demonstrated on synthetic and real-world datasets. Results are compared with some of the traditional clustering methods, showing that our method achieves similar or better results in terms of clustering accuracy.
\end{abstract}


\section{Introduction}
Data that is directional in nature and can be represented as unit vectors on a d-dimensional sphere has numerous interesting applications \cite{mardia1999directional, fisher1995statistical, pewsey2021recent}. Unit vectors are commonly used in wind data analysis \cite{lund1999cluster}. Meteorologists frequently collect wind direction and speed at multiple sites and utilize unit vectors to represent and analyse wind behaviour, as well as to statistically process and make future predictions about wind patterns. The field that regularly uses unit vectors is robotics, where unit vectors represent the orientation of robot parts and robots. For example, the orientation of a robotic arm can be represented using a unit vector, which leads to a simplified control of the arm’s movement \cite{zhang2022perspective}. Unit vectors are used in medicine as well, representing the orientation of limbs or joints during movement, enabling researchers to study and analyse the kinematics of human movement and the effects 
of various factors on it, such as injury or age \cite{rancourt2000using}.

In many of these applications, it is necessary to cluster directional data into distinct groups. Clustering is an unsupervised machine learning technique that divides data into different groups, ensuring that data points within a cluster are more similar to each other than to those in other clusters. Some common use cases of data clustering are image segmentation \cite{mittal2022comprehensive}, customer segmentation \cite{kansal2018customer}, gene expression analysis \cite{oyelade2016clustering}, and anomaly detection \cite{ariyaluran2022clustering}. Clustering is an important tool for 
discovering hidden patterns and data relationships.

Unit vectors can also be subjected to clustering, which involves grouping them based on their direction or orientation. There are several methods for clustering unit vectors, including k-means clustering \cite{mcqueen1967some} and its variant, spherical k-means \cite{ng2001spectral}. Spherical k-means is specifically designed for directional data clustering and uses cosine similarity as a distance measure instead of Euclidean distance. Spherical k-means has been used in medicine for breast cancer clustering \cite{rustam2020breast} or for acute sinusitis classification \cite{arfiani2019kernel}. The most common use of spherical k-means is text documents clustering \cite{dhillon2001concept}. Other popular techniques for grouping directional data are hierarchical clustering \cite{murtagh2012algorithms} and mixture models \cite{golzy2016algorithms}. Hierarchical clustering creates a hierarchy of clusters based on the similarity of data points. On the other hand, mixture models are statistical models, that assume that data is produced by a combination of some well-known distributions \cite{banerjee2003expectation, banerjee2005clustering, nguyen2017novel}. Density-based clustering is a different category of clustering algorithms, with DBSCAN (Density-Based Spatial Clustering of Applications with Noise) as a well-known example \cite{khan2014dbscan}. DBSCAN clusters data points according to their density in relation to other points. Overall, there are a variety of techniques for clustering directional data, but the best technique depends on the specific task at hand.

We have used synchronization phenomenon for clustering in this paper. Synchronization is when two or more oscillating systems match their phases over 
time. Clustering based synchronization is a method for grouping data points into 
clusters based on their degree of synchronization \cite{shao2012synchronization, bohm2010clustering, chen2014fast}. Highly synchronized data points can be grouped, whereas weakly synchronized data points can be divided into separate clusters. The use of this kind of clustering can help find patterns and connections in data that might not be obvious from other approaches. 

The Kuramoto model is the most common mathematical model (\ref{eq:kuramoto}) for studying synchronization phenomenon \cite{kuramoto2005self}. It describes the dynamics of $N$ coupled oscillators as:
\begin{equation}
\label{eq:kuramoto}
\frac{d\theta_i}{dt}=\omega_i+\frac{K}{N}\sum_{j=1}^{N}\sin(\theta_j-\theta_i), \quad (i=1,\dots,N)
\tag{1}
\end{equation}
where the oscillators are modelled by a system of coupled differential equations that describe the evolution of each oscillators’ phase $\theta_i$ over time. Their interaction is controlled by the coupling strength $K$, which controls how strong oscillators influence one another and determines the degree of synchronization in the system. Notation $\omega_i$ represents an intrinsic frequency of oscillator $i$. Variants of the Kuramoto model have been applied to the study of synchronization phenomenon in complex networks \cite{arenas2008synchronization}, data clustering \cite{crnkic2019data}, and rotation averaging and interpolation problems \cite{kapic2021new, kapic2023interpolation}. A generalized version of model (\ref{eq:kuramoto}) to higher dimensions is used in this paper. Each oscillator is represented as a unit vector $Q\in \mathbb{R}^d$, corresponding to a point on the $(d-1)$-dimensional unit hypersphere $\mathbb{S}^{d-1}$. The dynamics of a single oscillator in this generalized setting are governed by the differential equation:
\begin{equation}
\dot{Q}=WQ,
\tag{2}
\end{equation}
where $W$ is an antisymmetric $d\times d$ frequency matrix of the generalized oscillator. The system of coupled oscillators in this framework is described as:
\begin{equation}
\label{eq:hyperspherical_kuramoto}
\dot{Q}_{j,j=1,\dots,N} = \frac{K}{N}\sum_{i=1}^{N}\left(Q_i-\langle Q_j,Q_i\rangle Q_j\right)+W_jQ_j.
\tag{3}
\end{equation}
Equation (\ref{eq:hyperspherical_kuramoto}) extends the classical Kuramoto model (\ref{eq:kuramoto}) from the unit circle $\mathbb{S}^1$ to the unit hypersphere $\mathbb{S}^{d-1}$. To better understand the dynamics of the system (\ref{eq:hyperspherical_kuramoto}), we introduce order parameter:
\[
R=\frac{1}{N}\sum_{j=1}^{N}Q_j.
\]
Notice that $\|R\|$ (notation $\|\cdot\|$ stands for the Euclidean norm) is a real number and has a value between $0$ and $1$. Case $\|R\|=1$ corresponds to the system being a completely synchronized, $Q_i=Q_j$ for all $i,j$. On the other side, $\|R\|=0$ represents an incoherent state. In this work, we apply a variation of this generalized model for clustering tasks, taking advantage of its synchronization properties to cluster data on the unit hypersphere.

The remainder of the paper is organized as follows. In section 2, we introduce a novel approach for clustering data on the unit hyperspheres using a vector-based Kuramoto model. We also introduce a novel algorithm based on this approach. Section 3 demonstrates the performance of our algorithm through simulations and visualizations of both real-life and simulated high-dimensional circular data. Our approach is compared to some state-of-the-art algorithms. In the conclusion, we review our research and discuss possible future outlooks.

\section{Algorithm}

To introduce a clustering method for points $P_j, j=1,\dots,N$, belonging to the unit sphere $\mathbb{S}^{d-1}$, we consider a system of coupled differential equations that describe the evolution of the points under mutual influence. Each point follows a dynamical equation where its movement depends on the average position of all other points in the system. Setting the frequency matrix $W=0$, (\ref{eq:hyperspherical_kuramoto}) transforms to:
\begin{equation}
\dot{Q}_{j}, j=1,\dots,N = \frac{K}{N}\sum_{i=1}^{N}\left(Q_i-\langle Q_j,Q_i\rangle Q_j \right),
\tag{4}
\end{equation}
where $Q_j$ represents the position of each point on the hypersphere and $K$ is the coupling parameter, which we set to $K=1$ without loss of generality. The notation $\langle \cdot,\cdot \rangle$ stands for the standard dot-product in $\mathbb{R}^d$. Initial conditions are given as $Q_j(0)=P_j, j=1,\dots,N$. 

This model preserves the unit sphere $\mathbb{S}^{d-1}$, ensuring that the evolution of each $Q_j$ remains on the sphere for all time $t$. The dynamics of the system lead to the formation of clusters, where points with similar orientations group together over time. To extract the clusters from the evolved system, we analyse the pairwise cosine distances between the points obtained at time $t=T$. Two points $Q_i$ and $Q_j$ are considered to belong to the same cluster if their cosine distance is below a predefined threshold $\epsilon$.

Mathematically, it is expressed as
\[
d(Q_i,Q_j)=1-\frac{Q_i\cdot Q_j}{\|Q_i\|\|Q_j\|}<\epsilon.
\]

From this similarity measure, we construct an adjacency matrix $A$ where:
\[
A_{ij}=
\begin{cases}
1, & d(Q_i,Q_j)<\epsilon, \\
0, & \text{otherwise.}
\end{cases}
\]

The final clusters are extracted as connected components of the graph represented by $A$. We explain our clustering method in detail as follows:

\begin{itemize}
\item \textbf{Initialize}
\begin{itemize}
\item Input $N$ points $P_j\in\mathbb{S}^{d-1}, j=1,\dots,N$.
\item Set parameters: time step $\delta$, clustering threshold $\epsilon$, and $\nu$.
\end{itemize}

\item \textbf{Solve the dynamical system}
\begin{itemize}
\item Integrate the system of equations until stopping criteria $|(\|R(t+\delta)\|-\|R(t)\|)|<\nu$ is satisfied.
\item Take the obtained time $T=t$ and compute $Q_j(T)$.
\end{itemize}

\item \textbf{Construct adjacency matrix}
\begin{itemize}
\item Compute pairwise cosine distances between all points.
\item Define adjacency matrix $A$ based on threshold $\epsilon$.
\end{itemize}

\item \textbf{Extract clusters}
\begin{itemize}
\item Identify connected components of the graph represented by $A$.
\end{itemize}

\item \textbf{Return clusters}
\begin{itemize}
\item Output the set of clusters $C_1,C_2,\dots,C_k$ corresponding to groups of points with high mutual similarity.
\end{itemize}
\end{itemize}

This method enables the automatic clustering of points on the unit hypersphere based on their emergent dynamics. A classical fourth-order Runge-Kutta method is used for solving systems of ODE’s. All simulations were performed using R (version 4.4.2), employing built-in ODE solvers from the Runge-Kutta family. Visualizations were created using Wolfram Mathematica (version 13.1).

\section{Simulations}
In this section, we will evaluate the effectiveness of our clustering algorithm. To achieve this, we test its performance on various synthetic and real-world datasets and compare the results with the spkmeans (Spherical K-Means Clustering) \cite{hornik2012spherical} and movMF (Mixtures of von Mises-Fisher Distributions) \cite{hornik2014movmf} algorithms. These methods were chosen because they are widely used and specifically designed for hyperspherical data. Unlike spkmeans and movMF, our algorithm does not require the number of clusters to be specified in advance, making it practical for unsupervised settings. The experiments are implemented in R, and clustering effectiveness is measured using macro-recall, macro-precision, Normalized Mutual Information (NMI) and Adjusted Rand Index (ARI). Macro-recall represents an average recall over all classes and is calculated as:
\[
\text{Macro-recall}=\frac{1}{C}\sum_{i=1}^{C}\frac{TP_i}{TP_i+FN_i},
\]
where $C$ represents a number of classes, and $TP_i$ and $FN_i$ are true positives and false negatives for class $i$, respectively. Macro-precision represents an average precision over all classes:
\[
\text{Macro-precision}=\frac{1}{C}\sum_{i=1}^{C}\frac{TP_i}{TP_i+FP_i},
\]
where $FP_i$ is a false positive for class $i$. Macro-recall and macro precision metrics are useful when dealing with imbalanced datasets. While macro metrics focus on class-wise accuracy, NMI and ARI measure overall agreement between predicted clusters and true labels regardless of label order. Range for NMI is from 0 to 1, with 1 indicating perfect agreement. Range for ARI is from -1 to 1, where 1 means perfect match and 0 is random labelling.

\subsection*{3.1 Synthetic datasets}

We will utilize random data generated from the von Mises-Fischer distribution \cite{banerjee2005clustering}, a probability distribution over unit vectors on a sphere, to test the performance of our clustering algorithm. For these purposes, we have created two random datasets Dat\_1 and Dat\_2, one with 150 three-dimensional unit vectors, and the other one with 200 four-dimensional unit vectors.

In the dataset Dat\_1, we have three clusters of 50 data points generated from von Mises-Fischer distribution, where the first cluster has a mean direction of $\mu_1=(1,0,0)$, the second cluster has a mean direction of $\mu_2=(0,1,0)$, and the third cluster has a mean direction of $\mu_3=(0,0,1)$. The concentration parameter is fixed for all clusters and is set to $\kappa=20$.

Results of macro-recall, macro-precision, NMI and ARI for different clustering techniques over this dataset are given in Table 1. The proposed algorithm achieves the highest values across all metrics and clusters data into five clusters instead of the original three, with two of them being identified as outliers. The ability to detect and separate outliers demonstrates the algorithm’s sensitivity to distinct patterns and anomalies within the dataset. The proposed algorithm does not know the number of clusters in advance, while the other two require it to be specified at the start.

\begin{table}[h]
\label{tab:dat1}
\centering
\caption{Clustering report for the dataset Dat\_1}
\begin{tabular}{lccccc}
\hline
Algorithm & Macro-recall & Macro-precision & ARI & NMI & Number of clusters \\
\hline
spkmeans & 0.980 & 0.980 & 0.940 & 0.911 & 3 \\
movMF & 0.980 & 0.980 & 0.940 & 0.911 & 3 \\
The Algorithm & \textbf{0.987} & \textbf{0.987} & \textbf{0.960} & \textbf{0.942} & 5 \\
\hline
\end{tabular}
\end{table}

Fig. 1. illustrates the clustering results on synthetic dataset Dat\_1 sampled from $S^2$, obtained at time $T=1.27$.

\begin{figure}[H]
\centering
\includegraphics[width=0.6\textwidth]{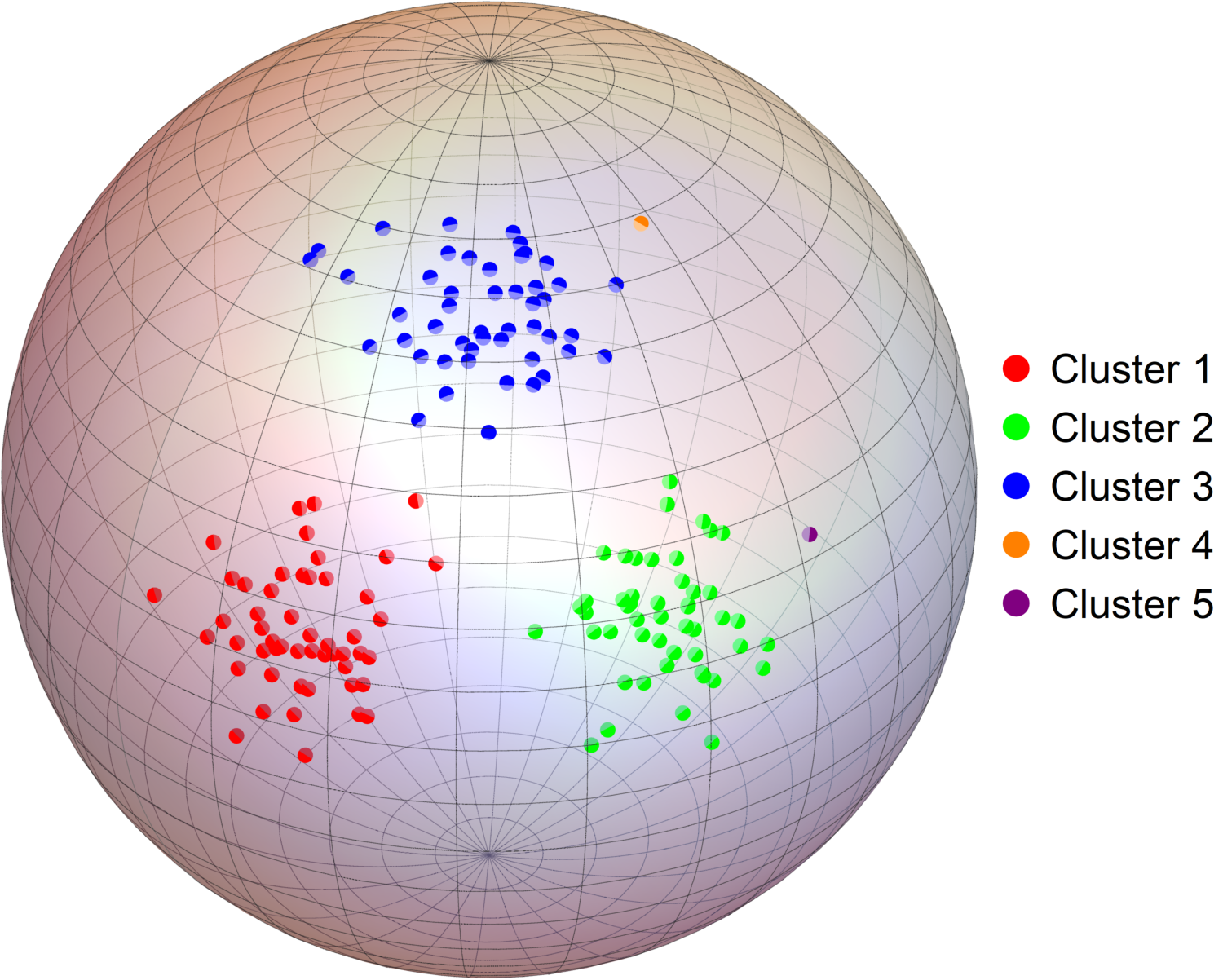}
\caption{Simulation results for the dataset Dat\_1 with clusters obtained at time $T=1.27$. The algorithm identified five clusters, including two (Cluster 4 and 5) corresponding to outliers.}
\end{figure}

Dat\_2 consists of 200 five-dimensional data points with 2 clusters generated with parameters $\mu_1=(1,0,0,0,0)$, $\mu_2=(-1,0,0,0,0)$, and $\kappa=20$. The data points lie in five-dimensional space and therefore cannot be directly visualized.

Fig. 2. illustrates the order parameter over time, showing the synchronization process. When the order parameter reaches 1, full synchronization occurs. We select a moment before this point to examine the number of clusters. Results presented in Table 2 are obtained at moment $T = 4.6$.

\begin{figure}[H]
\centering
\includegraphics[width=0.6\textwidth]{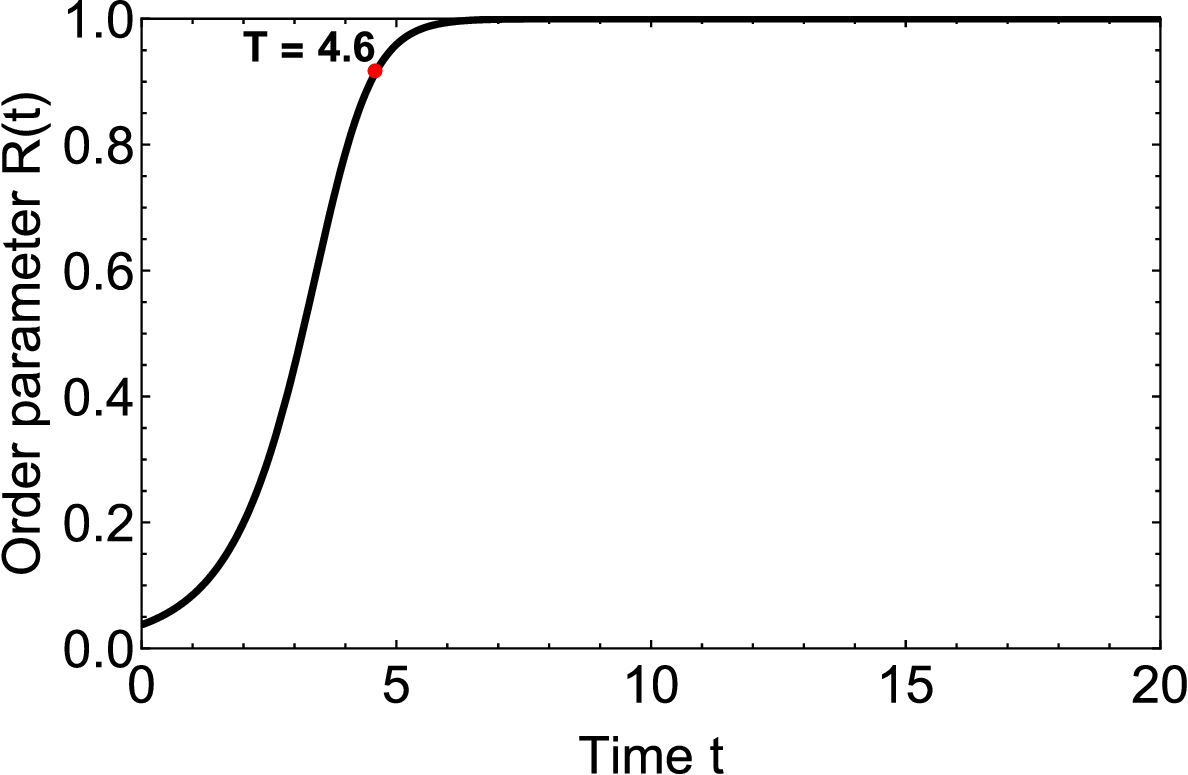}
\caption{Order parameter during the synchronization process for the dataset Dat\_2}
\end{figure}

The moment $T$ is chosen just before full synchronization occurs. That is a moment when the order parameter stops changing much. At that moment we find meaningful clusters. This approach is similar to the hierarchical clustering and cutting a dendrogram at the right level to find meaningful clusters before everything merges into one. Table 2 presents the macro-recall, macro-precision, NMI, and ARI results for the algorithms on the second dataset.

\begin{table}[h]
\centering
\caption{Clustering report for the dataset Dat\_2}
\begin{tabular}{lccccc}
\hline
Algorithm & Macro-recall & Macro-precision & ARI & NMI & Number of clusters \\
\hline
spkmeans & 0.995 & 0.995 & 0.980 & 0.959 & 2 \\
movMF & \textbf{1.000} & \textbf{1.000} & \textbf{1.000} & \textbf{1.000} & 2 \\
The Algorithm & 0.980 & 0.995 & 0.980 & 0.959 & 2 \\
\hline
\end{tabular}
\end{table}

The algorithm has demonstrated its applicability to higher dimensions. In the case of five-dimensional unit vectors, it achieved macro-precision, macro-recall, NMI, and ARI results that are competitive with the compared algorithms.

\subsection*{3.2 Real-world datasets}

We can evaluate our method using real-world datasets. Real-world datasets used are household expenditure survey \cite{hothorn2009handbook} and Iris dataset \cite{fisher1936use}. The data points are projected onto the sphere by scaling them to have unit length.

Household dataset is obtained from the R software and “HSAUR2” package. This dataset is a survey on household expenditure that has data separated into 2 clusters, men and women based on four parameters: housing, food, goods, and services. This dataset has 40 data points belonging to a dimension $S^3$. Each of these $\mathbb{R}^4$ vectors are normalized to have a unit length of 1. Table 3 presents the macro-recall, macro-precision, NMI, and ARI results for the algorithms on the Household dataset.

\begin{table}[h]
\centering
\caption{Clustering report for the Household dataset}
\begin{tabular}{lccccc}
\hline
Algorithm & Macro-recall & Macro-precision & ARI & NMI & Number of clusters \\
\hline
spkmeans & 0.825 & 0.847 & 0.408 & 0.371 & 2 \\
movMF & 0.825 & 0.870 & 0.409 & 0.443 & 2 \\
The Algorithm & \textbf{0.850} & \textbf{0.885} & \textbf{0.478} & \textbf{0.510} & 2 \\
\hline
\end{tabular}
\end{table}

Our algorithm outperformed both spkmeans and movMF algorithms in all evaluated metrics.

Iris dataset has measurements of 150 iris flowers from three different species: setosa, versicolor, and virginica. The dimensionality of this dataset is 4 and its data is represented as points on $S^3$. Table 4 summarizes the macro-recall, macro-precision, NMI, and ARI results on the Iris dataset.

\begin{table}[h]
\centering
\caption{Clustering report for the Iris dataset}
\begin{tabular}{lccccc}
\hline
Algorithm & Macro-recall & Macro-precision & ARI & NMI & Number of clusters \\
\hline
spkmeans & 0.967 & \textbf{0.969} & \textbf{0.904} & \textbf{0.898} & 3 \\
movMF & 0.953 & 0.959 & 0.868 & 0.871 & 3 \\
The Algorithm & \textbf{1.0} & 0.667 & 0.568 & 0.734 & 2 \\
\hline
\end{tabular}
\end{table}

The method identifies two clusters: one that precisely matches Iris setosa, while the other combines Iris virginica and Iris versicolor. This result aligns with expectations in unsupervised learning, as these two species cannot be easily distinguished without category labels. Both the spkmeans and movMF algorithms showed potential instability, as different runs with varying random seeds produced different results. Our simulations confirmed this behaviour, suggesting sensitivity to initialization. In contrast, our algorithm produced consistent clustering outcomes across multiple runs.

\section{Conclusion}
In this paper, we have introduced a new algorithm for performing directional data clustering. This algorithm belongs to the group of synchronization clustering algorithms because it is based on the extension of the classical Kuramoto model to the unit hypersphere. We demonstrated the effectiveness of our algorithm on both real-world and synthetic datasets, showing that it achieves comparable or superior clustering accuracy compared to established techniques such as spherical k-means and the movMF algorithm. The algorithm showed effectiveness in identifying outliers inside data. Given that the algorithm does not require defining the number of clusters in advance and autonomously discovers the structure of groups, it belongs to the category of unsupervised algorithms. This feature makes it more practical in real problems where in most cases the number of groups is not known in advance. However, its reliance on numerically solving differential equations introduces computational cost, especially for large datasets. As part of future work, we plan to extend the evaluation to larger datasets, improve computational cost, and investigate the performance on other non-Euclidean manifolds.

\bibliographystyle{unsrt}
\bibliography{main}
\end{document}